\documentclass[11pt]{article}

\usepackage[hyperref]{ccl2025-en}
\hypersetup{hypertexnames=false}
\makeatletter
\let\cclOriginalLbibitem\@lbibitem
\def\@lbibitem[#1]#2{%
  \cclOriginalLbibitem[#1]{#2}%
  \Hy@raisedlink{\hyper@anchorstart{cite.#2}\hyper@anchorend}}
\makeatother
\usepackage{times}
\usepackage{url}
\usepackage{latexsym}
\usepackage{fancyhdr}
\usepackage[T1]{fontenc}
\usepackage[utf8]{inputenc}
\usepackage{microtype}
\usepackage{inconsolata}
\usepackage{graphicx}
\usepackage{booktabs}
\usepackage{multirow}
\usepackage{array}
\usepackage{amsmath}
\usepackage{amssymb}
\usepackage{float}
\usepackage{flafter}
\usepackage{placeins}

\graphicspath{{figures/}}

\newcommand{\method}{\textsc{RVV}}
\title{Route--Verify--Vote: Procedure-Conditioned Self-Consistency for Mixed-Domain Reasoning}

\author{Xinchen Xiao \\
  Xinjiang University \\
  \texttt{xiaoxinchenn@outlook.com}}
\date{}

\begin{document}
\maketitle
\setcounter{footnote}{0}

\begin{abstract}
Compositional generalization remains challenging when language models must combine familiar reasoning operations in unfamiliar ways. The Scenario-Based Commonsense Reasoning Evaluation (SCoRE) 2026 tests this ability on three mixed domains absent from training and requires models to identify the complete set of correct options for each question.

We introduce Route--Verify--Vote (\method), a framework for procedure-conditioned self-consistency that uses language models without parameter updates. Route uses the provided domain label to select a reasoning procedure that guides the model in representing and applying the relevant constraints. Verify prompts the model to assess each option against those constraints. Vote aggregates complete answer sets and allocates additional samples to questions with a small vote-count margin between the two most frequent sets. Samples for each question follow the same domain-specific procedure.

On the official test set, voting over 16 sampled answer sets per question achieves an exact-set accuracy of 74.6\%. Adaptive \method{} reaches 77.3\%, and combining models on selected domain routes raises accuracy to 79.4\%. The final system ranked second among participating systems. These results support domain-specific reasoning procedures and answer-set disagreement as useful tools for allocating inference-time computation in mixed-domain reasoning.
\englishkeywords{mixed-domain reasoning, procedure-conditioned
self-consistency, exact-set reasoning, test-time compute scaling, compositional
generalization, large language models}
\end{abstract}

\section{Introduction}

Mixed-domain reasoning is difficult not only because individual constraints
are hard, but because familiar operations must be recomposed under a changed
test distribution.  SCoRE 2026 isolates this challenge
\cite{zhan2026score}.  Training
contains four atomic domains and only one mixed domain; testing introduces
three unseen mixed domains and requires an exact subset of correct options.
The solver must transfer both primitive operations and their ordering.

Our early experiments revealed a consistent train--test difference.  Qwen3.5-9B
reached 90.0\% majority accuracy on a 50-example training subset but only
35.9\% on the official test set under a smaller sampling budget.  The
Qwen-3.5-28B-A3B-REAP checkpoint\footnote{Model repository: \url{https://huggingface.co/0xSero/Qwen3.5-28B}.} (hereafter REAP) reached
82.33\% on a stratified
300-example training subset, whereas its available official-test run scores
averaged $70.67\pm0.81$\%.  Because the protocols differ, we treat these
comparisons as descriptive.  This consistent pattern,
together with the unseen domain compositions, suggested that evaluations on training subsets
were a poor proxy for official-test difficulty.

Our preliminary dense-pruning, fine-tuning, and reinforcement-learning
experiments did not yield consistent improvements on the official test set.  This observation aligns with prior findings that parameter updates
can distort pretrained features under distribution shift
\cite{kumar2022finetuning}.  We therefore kept the model frozen and shifted
adaptation to its inference procedure and compute budget.

Inference-time computation can allocate effort according to question
difficulty \cite{snell2024testtime}, but unconstrained sampling alone is
insufficient here.  A solver must choose a state representation, compose
operators in the correct order, and inspect every option.  Generic
chain-of-thought prompting can elicit intermediate steps
\cite{kojima2022zeroshot}, but does not enforce these procedural choices or a
complete exact-set decision.

Procedural control does not eliminate sampling variability.  Individual
runs differ on 179--248 of 1,000 test questions even though each produces
parseable non-empty answer sets for 986--991 questions.  Because parsing
coverage is nearly constant,
this variation calls for aggregation over complete answer sets rather than
reliance on a single trace.

We call the resulting abstraction \emph{procedure-conditioned
self-consistency}.  Each route supplies the same fixed natural-language
procedure to all samples, specifying a state representation, operators,
inference order, consistency check, and output format.  Parseable generations
are mapped to the same complete-answer-set space, so
their agreement can serve as a compute-allocation signal.

\method{} instantiates this abstraction in three stages.  \emph{Route} selects
one of eight procedures; the frozen language model then constructs a state such
as temporal equations, spatial coordinates, or a social relation graph.
\emph{Verify} requests a separate judgment for all four options against that
state.  \emph{Vote} aggregates parseable answer sets and targets extra samples
to low-margin questions.  The final system extends the core
REAP-only \method{} with route-conditioned model pairing before the final vote.

Our contribution is a system-level study of this abstraction under mixed-domain
exact-set evaluation.  We describe how prompted route-specific state
construction, complete-set verification, and margin-guided sampling form one
inference-time control loop. The final system achieves 79.4\% accuracy and ranks second.

\section{Task and Evaluation}

Each instance contains a context $x$, question $q$, four options
$O=\{o_A,o_B,o_C,o_D\}$, a language label, and a domain label.  The target is
a non-empty subset $Y\subseteq\{A,B,C,D\}$.  We evaluate exact set match:
an answer is correct only when every selected and omitted option matches the
gold set, following the official task specification and accuracy metric
\cite{zhan2026score,score2026task}.  Table~\ref{tab:data} reports the released split by
domain.

\begin{table}[!htbp]
  \centering
  \small
  \begin{tabular}{lrr}
    \toprule
    Domain & Train & Test \\
    \midrule
    Nature & 1,000 & 100 \\
    Social & 500 & 25 \\
    Space & 1,000 & 100 \\
    Time & 1,000 & 100 \\
    \midrule
    Space+Nature & 100 & 75 \\
    Space+Social & 0 & 200 \\
    Time+Nature & 0 & 200 \\
    Time+Social & 0 & 200 \\
    \midrule
    Total & 3,600 & 1,000 \\
    \bottomrule
  \end{tabular}
  \caption{Released SCoRE 2026 data by domain.  Three mixed-domain labels in
  the test set do not occur in the released training set.}
  \label{tab:data}
\end{table}

The training set contains 2,100 Chinese and 1,500 English questions; the test
set contains 550 Chinese and 450 English questions.  Multiple-correct-option
cases are substantial rather than exceptional: 848 of the 3,600 training
questions have two or more correct options.  In combination, these properties
create three difficulties.
First, constraints must be normalized into a representation suited to their
domain.  Second, mixed-domain questions require ordered composition: resolving
an entity incorrectly before spatial deduction invalidates the remaining
trace.  Third, exact-set evaluation penalizes premature stopping and
over-selection equally.

\section{Related Work}

\paragraph{Prompted reasoning.}
Chain-of-thought prompting elicits intermediate steps for arithmetic,
commonsense, and symbolic tasks \cite{wei2022cot}; zero-shot variants show
that a short reasoning instruction can also improve diverse tasks
\cite{kojima2022zeroshot}.  Least-to-most prompting decomposes difficult
problems into sequential subproblems \cite{zhou2022leasttomost}, while Program
of Thoughts delegates computation to executable programs
\cite{chen2022pot}.  In contrast, our system selects a predefined state
representation and verification procedure tailored to each domain, with an
explicit operation order for mixed-domain questions.

\paragraph{Reasoning-path aggregation.}
Self-consistency samples multiple reasoning paths and marginalizes them at the
answer level \cite{wang2022selfconsistency}.  We apply this principle to
exact-set outputs generated under a shared route-specific prompt.  This
procedure-conditioned view supplies the same requested representation and
inference order across samples, aggregates complete answer sets, and retains
the vote distribution as a compute-allocation signal. A prior CCL evaluation
system report also combines self-retrieval-augmented generation with
multi-round accumulative voting, for fine-grained Chinese hate-speech
recognition \cite{wang-etal-2025-srag-mav}; our study instead examines
procedure-conditioned exact-set reasoning and margin-guided allocation on
mixed-domain questions.

Test-time compute can be more effective when it is allocated according to
instance difficulty rather than uniformly \cite{snell2024testtime}.  \method{}
combines answer-set margin with procedure conditioning in a frozen-model
inference loop.

\paragraph{Compositional generalization.}
Compositional evaluation asks whether familiar primitives remain usable in new
combinations.  Existing work finds that accuracy decreases as test compounds
diverge from training compounds \cite{keysers2019compositional}.  SCoRE offers
a practical instance of this setting: three mixed-domain labels occur only in
the test set and must be solved by composing familiar domain procedures.

\section{Procedure-Conditioned Self-Consistency Framework}

\begin{figure}[t]
  \centering
  \includegraphics[width=\textwidth]{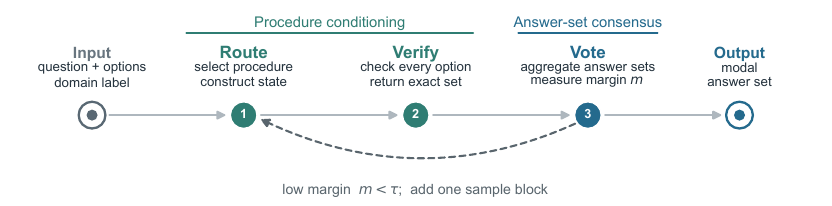}
  \caption{Procedure-conditioned view of \method{}.  Routing selects a fixed
  prompt that requests a domain-specific representation and inference order;
  option-wise verification produces a complete answer set when parsing
  succeeds; and voting aggregates repeated samples.  Low-margin questions
  receive additional samples under the same procedure prompt.}
  \label{fig:overview}
\end{figure}

Figure~\ref{fig:overview} illustrates the unified \method{} architecture.  All eight
domain labels share the same route--verify--vote control flow, while each route
defines its own constraint state and core operators.
Table~\ref{tab:route-operations} summarizes the route families.

\begin{table}[t]
  \centering
  \footnotesize
  \setlength{\tabcolsep}{4pt}
  \begin{tabular}{@{}l>{\raggedright\arraybackslash}p{0.29\textwidth}
    >{\raggedright\arraybackslash}p{0.43\textwidth}@{}}
    \toprule
    Route & Prompted state & Requested operation order ($\prec_d$) \\
    \midrule
    Nature & attribute table & normalize attributes; check fact consistency \\
    Social & relation graph & resolve entities; traverse relation paths \\
    Space & coordinates or seats & fix reference frame; apply offsets and exclusion \\
    Time & event variables & normalize events; solve relative or modular equations \\
    \midrule
    Space+Nature & attributes and coordinates & resolve attributes; substitute entities; solve layout \\
    Space+Social & relation graph and coordinates & resolve social references; substitute entities; solve layout \\
    Time+Nature & attributes and event variables & resolve attributes; substitute events; solve timeline \\
    Time+Social & relation graph and event variables & resolve social references; substitute events; solve timeline \\
    \bottomrule
  \end{tabular}
  \caption{The eight route templates and their specified reasoning procedures. Every
  prompt also requests a full-constraint recheck, option-wise judgments, and
  the same JSON output format.}
  \label{tab:route-operations}
\end{table}

\subsection{Route: Select a Procedure}

Let $d$ be the provided domain label.  A deterministic lookup maps it to a
prompt $P_d$ from eight fixed templates:
\begin{equation}
  P_d = \mathcal{P}[d], \qquad
  d\in\{n,s,p,t,pn,ps,tn,ts\}.
\end{equation}
Here $n,s,p,t$ denote Nature, Social, Space (using $p$ for position to avoid an
abbreviation conflict), and Time.  Concatenated symbols denote mixed domains;
for example, $pn$ and $ps$ denote Space+Nature and Space+Social.
Routing changes the requested reasoning procedure, not merely the
persona.  Every route prompt specifies (i) the state representation, (ii) operators
for translating linguistic constraints, (iii) an inference order, (iv) a
global consistency check, and (v) a structured output format.
We write this operational prompt specification as
\begin{equation}
  \Pi_d=(\mathcal{S}_d,\mathcal{O}_d,\prec_d,\mathcal{V},\mathcal{C}),
\end{equation}
where $\mathcal{S}_d$ is the requested route-specific state,
$\mathcal{O}_d$ its requested operators, $\prec_d$ their requested execution
order, $\mathcal{V}$ the full-constraint and option-wise check, and
$\mathcal{C}$ the exact-set output format.
In these templates, state and operator instructions express
$\mathcal{S}_d$ and $\mathcal{O}_d$; ordered reasoning steps express $\prec_d$;
full-constraint and option-wise checks express $\mathcal{V}$; and a JSON answer
object realizes $\mathcal{C}$. The template $P_d$ specifies the reasoning
procedure through natural-language instructions; programmatic checks apply
only to the output format $\mathcal{C}$.

\subsection{Construct a Route-Specific Constraint State}

The nature route organizes taxonomic and attribute evidence before evaluating a
claim.  The social route constructs a directed relation graph and resolves
descriptions such as kinship chains to named entities.  The spatial route maps
entities to coordinates or seats and applies mutual-exclusion constraints.  The
temporal route maps each event $e$ to a time variable $T(e)$ and translates
relative expressions into equations or modular offsets.

Mixed-route prompts compose these primitives in an explicit order.  For example, the
space+social route first resolves relational descriptions into entities and
then substitutes those entities into a coordinate-based constraint system.
This ordering prevents unresolved descriptions from being treated as distinct
spatial objects.  The design resembles decomposition methods that solve simpler
subproblems before the original problem \cite{zhou2022leasttomost}, but requests
a domain-specific state transition rather than asking the model to invent a
decomposition for each example.

\subsection{Verify: Predict a Complete Answer Set}

Within the $k$th generation $z_k$, the frozen model is prompted to judge every
option $o_j$ separately and then serialize all options judged true into one
JSON answer list.  Let $v_j^{(k)}$ denote the corresponding judgment within the generated response:
\begin{equation}
  \hat{y}_{j}^{(k)} = \mathbb{I}[v_j^{(k)}=\textsc{true}]\in\{0,1\},
  \quad j\in\{A,B,C,D\}.
\end{equation}
Here $v_j^{(k)}\in\{\textsc{true},\textsc{false}\}$ is the prompted Boolean
judgment based on the derived state $S^{(k)}$, and
$\hat{Y}^{(k)}=\{j:\hat{y}_{j}^{(k)}=1\}$.  The route prompt requests a final
constraint recheck before serialization. Verify is a self-check within the
same generation, and the parser extracts the final answer list.

\subsection{Vote: Aggregate Answer Sets}

Long reasoning traces are sensitive to sampling.  We generate $K$ stochastic
solutions with fixed decoding parameters, parse each output, and vote over
parseable non-empty answer sets:
\begin{equation}
  \begin{aligned}
  \mathcal{Y}_{\mathrm{valid}}
    &=2^{\{A,B,C,D\}}\setminus\{\emptyset\},\\
  \hat{Y}
    &=\arg\max_{Y\in\mathcal{Y}_{\mathrm{valid}}}\sum_{k=1}^{K}
      \mathbb{I}[\operatorname{Parse}(z_k)=Y].
  \end{aligned}
\end{equation}
We define the top-two margin $m$ as the difference between the vote counts of
the most frequent and second-most frequent answer sets.  When only one answer
set is observed, the runner-up count is zero.
Adaptive sampling continues while $m<\tau$.  The final system uses
$\tau=12$ at 32-sample increments.
This answer-level aggregation follows the self-consistency principle of
sampling diverse reasoning paths and selecting their most consistent answer
\cite{wang2022selfconsistency}.  Empty or invalid parses are recorded but do
not become answer candidates when at least one parseable non-empty set exists; an
all-invalid question remains unresolved.  The final configuration uses
temperature 0.85, top-$p$ 1.0, min-$p$ filtering at 0.05, and up to 36,000 generated
tokens.  Exact modal ties are resolved by choosing the answer whose supporting
generations have the shorter mean length.

\paragraph{Final-system pairing.}
Core \method{} uses REAP samples only.  The final configuration
adds \emph{route-conditioned model pairing}: after adaptive REAP sampling fixes
the per-question budget, Nature, Space, and Space+Social receive the same
number of samples from Qwen3-30B-A3B-Thinking-2507.  The two answer-set pools
are merged before the same Vote stage.  Pairing therefore inherits the
per-question cap of 512 samples per model.  The three paired routes were fixed
using domain-wise official test feedback.

\section{Official Test Results and Agreement Analysis}

\subsection{Experimental Setup}

The core model is Qwen-3.5-28B-A3B-REAP, served locally with vLLM 0.19.1 in
bfloat16 without quantization; \emph{REAP} denotes this checkpoint.  The final
configuration also invokes API-served Qwen3-30B-A3B-Thinking-2507 on Nature,
Space, and Space+Social through route-conditioned model pairing.

Intermediate scores were provided by the organizers, who also reproduced
the final system's accuracy of 79.4\%.  One of 16 early runs lacks an individual score; it
remains in the pooled vote but is omitted from score summaries.  Self-hosted
REAP experiments used about 280 H800 GPU-hours.
Table~\ref{tab:voting} summarizes accuracy at each inference stage.

\begin{table}[!htbp]
  \centering
  \small
  \setlength{\tabcolsep}{4pt}
  \begin{tabular}{@{}lcc@{}}
    \toprule
    Inference method & Evaluation setting & Accuracy (\%) \\
    \midrule
    Individual, mean $\pm$ SD & 15 scored runs & $70.67\pm0.81$ \\
    Individual range & 15 scored runs & 68.9--72.3 \\
    \midrule
    Modal answer vote & 16 pooled outputs & 74.6 \\
    Uniform \method{} & REAP, 32 outputs & 75.9 \\
    Adaptive \method{} & REAP, up to 512 & 77.3 \\
    Final system & \method{} + route-conditioned pairing & \textbf{79.4} \\
    \bottomrule
  \end{tabular}
  \caption{Official-test exact-set accuracy.  The pooled vote uses all 16
  outputs; score summaries omit the unavailable individual result.  Adaptive
  REAP-only \method{} reaches 77.3\%, and model pairing adds 2.1 points.}
  \label{tab:voting}
\end{table}

\subsection{Comparison of Domain-Specific Configurations}

Table~\ref{tab:domain-prompt-ablation} compares initial and selected
configurations.  Test-size-weighted accuracy rose from 35.9\% to 56.0\%; the
largest domain difference is 44.0 points on Space+Social.

\begin{table}[!htbp]
  \centering
  \small
  \setlength{\tabcolsep}{6pt}
  \begin{tabular}{@{}lrrr@{}}
    \toprule
    Domain & Initial baseline & Selected configuration & $\Delta$ pp \\
    \midrule
    Nature & 23.0 & 44.0 & +21.0 \\
    Social & 24.0 & 28.0 & +4.0 \\
    Space & 23.0 & 41.0 & +18.0 \\
    Time & 34.0 & 39.0 & +5.0 \\
    \midrule
    Space+Nature & 38.7 & 56.0 & +17.3 \\
    Space+Social & 36.5 & \textbf{80.5} & \textbf{+44.0} \\
    Time+Nature & 49.5 & 68.0 & +18.5 \\
    Time+Social & 36.0 & 45.0 & +9.0 \\
    \midrule
    Test-size-weighted total & 35.9 & 56.0 & +20.1 \\
    \bottomrule
  \end{tabular}
  \caption{Comparison of the initial shared-prompt baseline and selected
  domain-specific configurations, using results from separate domain evaluations.
  Configurations can differ in model, prompt, and decoding settings.
  $\Delta$ denotes a
  percentage-point change.}
  \label{tab:domain-prompt-ablation}
\end{table}

Domain-specific scores are non-monotonic: Space+Social falls to 35.0\% before
reaching 80.5\%, while Time+Social falls from 45.0\% to 34.5\%.
Appendix Tables~\ref{tab:model-ablation}--\ref{tab:prompt-trajectory}
and Figure~\ref{fig:sensitivity} separately report model, early-prompt,
and decoding analyses.

\subsection{Voting, Adaptive Allocation, and Final Result}

The pooled 16-run vote reaches 74.6\%, versus a $70.67\pm0.81$\% mean and
68.9--72.3\% range across scored runs.  Its vote distribution is then reused
to allocate subsequent computation.

Adaptive \method{} adds REAP samples in blocks of 32 while $m<12$, up to 512,
and reaches 77.3\%.  Equal-count Qwen3-30B-A3B-Thinking-2507
samples on three routes raise the merged vote to 79.4\% (Appendix
Figure~\ref{fig:test-scaling}).

\subsection{Agreement Structure}

Across 16 outputs, 457 questions are unanimous.  Pairwise answer distance, the
number of questions on which two runs differ, ranges from 179 to 248 (mean
214.3), and 90 questions have margin at most two.  Each run yields parseable
non-empty sets for 986--991 questions, so parse failure alone cannot explain the
variation; six questions remain unresolved after pooling.

Adding 16 samples to the 201 questions with margin at most six changes 33
answers, including 29 in the margin-0--2 bins.  This concentration suggests that a small vote margin identifies
unstable predictions.  Appendix Figures~\ref{fig:run-variability}
and~\ref{fig:vote-allocation} show run variation and change rates; correction
and damage denote wrong-to-correct and correct-to-wrong changes, respectively.

\section{Discussion and Conclusion}

The system follows a two-step strategy: select a domain-specific procedure
before sampling, then allocate additional samples based on answer-set disagreement.  Once
prompts are fixed, pooled voting improves over individual runs and identifies
questions with small vote margins.  Uniform voting adds
only 0.3 points from $K=128$ to $K=512$, favoring targeted allocation here.

Adaptive \method{} reaches 77.3\%; route-conditioned model pairing adds 2.1
points, yielding 79.4\% accuracy and second place. Given the domain
labels and fixed prompts, complete-set disagreement offers a useful
signal for allocating test-time computation.

\section*{Limitations}

Leaderboard feedback informed prompts and pairing, so the official score
evaluates the final system rather than isolated components.  Routing assumes
provided domain labels.  The stopping threshold of 12 depends on the sampling schedule and does
not maintain a constant confidence level across budgets; a normalized
stopping criterion may be more efficient.

\noindent\begin{minipage}{\linewidth}
\section*{Acknowledgements}

The author is grateful to two friends who wish to remain anonymous: one for generously providing the computing resources and infrastructure needed for the experiments, and the other for their continued care, encouragement, and emotional support throughout this work. Without their help and encouragement, this work would have been much more difficult to complete.

The author also thanks 0xSero for releasing the REAP-pruned Qwen3.5-28B checkpoint used in this study, the Qwen team for the underlying models, and the researchers behind REAP for their work on expert pruning. Their publicly available models and methods made the inference experiments in this paper possible. The author thanks the SCoRE 2026 organizers for providing the benchmark and conducting the evaluation.

\par
\end{minipage}

\appendix
\raggedbottom

\section*{\LARGE Appendix}

The appendix provides supplementary implementation details and analyses: Section A describes the implementation, early train--test comparisons, and answer-set cardinality; Section B presents additional analyses of models, prompts, decoding, test-time scaling, and sampling variability and allocation.

\section{Implementation Details}
\label{sec:competition-refinements}
\small

The final configuration uses temperature 0.85, top-$p$ 1.0, min-$p$
filtering at 0.05, disabled top-$k$ filtering, a 40,000-token context, and at
most 36,000 generated tokens.
The locally hosted REAP environment uses Python 3.12.3, PyTorch 2.10.0 with CUDA 12.8,
Transformers 5.6.0, vLLM 0.19.1 with its PagedAttention serving architecture
\cite{kwon2023vllm}, and NVIDIA H800 PCIe GPUs. Total compute across self-hosted
REAP experiments, including exploratory and unsuccessful runs, was approximately
280 H800 GPU-hours. Compute for models accessed through external APIs is
excluded because provider-side GPU time is unavailable.

The final system computes the stopping margin on the REAP pool,
starting with 32 samples and adding blocks of 32 below margin 12, up to 512.
Once this budget is fixed, Nature, Space, and Space+Social receive an equal
number of Qwen3-30B-A3B-Thinking-2507 samples.  The final vote merges both
answer-set pools; all other routes remain REAP-only. Sample allocation and
results by domain are reported in Section B.2.

\subsection{Early Train--Test Comparisons}

Table~\ref{tab:train-test-gap} summarizes the observations that motivated the
shift away from parameter updates. Accuracy was consistently higher on training
subsets, making these evaluations poor guides to official-test difficulty.

\begin{table}[!htbp]
  \centering
  \scriptsize
  \setlength{\tabcolsep}{5pt}
  \begin{tabular}{@{}llrrl@{}}
    \toprule
    Model & Evaluation set & $N$ & $K$ & Accuracy (\%) \\
    \midrule
    Qwen3.5-9B & 50-example train subset & 50 & 32 & 90.0 \\
    Qwen3.5-9B & Official test & 1,000 & 8 & 35.9 \\
    \midrule
    Qwen-3.5-28B-A3B-REAP & 300-example train subset & 300 & 1 & 82.33 \\
    Qwen-3.5-28B-A3B-REAP & Official test & 1,000 & 1 & $70.67\pm0.81$ \\
    \bottomrule
  \end{tabular}
  \caption{Preliminary training-subset and test-set results. Prompts and
  sampling budgets differ across evaluations.}
  \label{tab:train-test-gap}
\end{table}

Preliminary dense 9B pruning provided no accuracy or speed advantage, while
fine-tuning and reinforcement learning showed no consistent gains. Incomplete
records limit these observations to a qualitative summary.

\subsection{Answer-Set Cardinality}

The released training labels contain 848 multi-answer questions.
Table~\ref{tab:answer-cardinality} shows that these include both two-option and
higher-cardinality targets, making exact-set verification relevant beyond a
small exceptional subset.

\begin{table}[!htbp]
  \centering
  \scriptsize
  \setlength{\tabcolsep}{8pt}
  \begin{tabular}{@{}lrrrr@{}}
    \toprule
    Correct options & 1 & 2 & 3 & 4 \\
    \midrule
    Questions & 2,752 & 440 & 304 & 104 \\
    Share (\%) & 76.4 & 12.2 & 8.4 & 2.9 \\
    \bottomrule
  \end{tabular}
  \caption{Answer-set cardinality in the 3,600-question released training set.
  Counts are computed directly from the gold answer lists.}
  \label{tab:answer-cardinality}
\end{table}

\section{Additional Analyses}

\subsection{Model, Prompt, and Decoding Comparisons}

Table~\ref{tab:model-ablation} compares three checkpoints under the same
50-example protocol.  The selected REAP checkpoint matches
Qwen3-30B-A3B-Thinking-2507 while
running faster in this evaluation. The newer REAP variant has lower accuracy, parse
rate, and completion stability.

\begin{table}[!htbp]
  \centering
  \scriptsize
  \setlength{\tabcolsep}{4pt}
  \begin{tabular}{@{}lrrrrr@{}}
    \toprule
    Checkpoint & Acc. & Parse & Length stop & Empty & Time (s) \\
    \midrule
    Qwen-3.5-28B-A3B-REAP & \textbf{88.0} & \textbf{98.0} & 0 & 1 & 456.98 \\
    Qwen3-30B-A3B-Thinking-2507 & \textbf{88.0} & \textbf{98.0} & 1 & 1 & 536.28 \\
    Qwen3.6-28B-A3B-REAP & 74.0 & 88.0 & 7 & 6 & \textbf{443.75} \\
    \bottomrule
  \end{tabular}
  \caption{Same-protocol model ablation on 50 questions: P2, one sample,
  temperature 0.9, top-$p$ 1.0, min-$p$ filtering at 0.05, and a 22,000-token
  generation limit.  Accuracy and parse rate are percentages.  Length stop
  counts generations terminated at the token limit; Empty counts outputs
  without a parsed answer.}
  \label{tab:model-ablation}
\end{table}

Table~\ref{tab:generic-prompt-ablation} compares five prompts in a preliminary
experiment. P1 uses a general step-by-step reasoning procedure; P2 strengthens entity
alignment, reference-frame checks, and option-wise decisions; P3 uses
option-driven contradiction tests; P4 uses grid/CSP elimination; and P5 uses
formal equations and coordinate models.  With
$N=50$, one answer changes vote accuracy by two points.  P2 exceeds P5 by four
points despite nearly identical mean output lengths, while P3 produces shorter outputs than
P1 and achieves four percentage points higher voting accuracy.  These results rank the five complete
prompts in this evaluation; because several instructions differ at
once, they do not isolate a single prompting mechanism.

\begin{table}[H]
  \centering
  \scriptsize
  \setlength{\tabcolsep}{6pt}
  \begin{tabular}{@{}lrrrrr@{}}
    \toprule
    Prompt & Requests & Parse & Request acc. & Vote acc. & Tokens \\
    \midrule
    P1 & 350 & 94.86 & 82.57 & 86.0 & 9,059 \\
    P2 & 350 & 95.14 & \textbf{86.57} & \textbf{92.0} & 9,366 \\
    P3 & 300 & \textbf{96.67} & 84.00 & 90.0 & 8,988 \\
    P4 & 300 & 96.33 & 84.33 & 86.0 & \textbf{8,567} \\
    P5 & 300 & 95.33 & 84.67 & 88.0 & 9,340 \\
    \bottomrule
  \end{tabular}
  \caption{Early Qwen3.5-9B prompt ablation on the same 50 questions.  The
  experiment contains 1,600 requests and 32 total samples per question across
  the five prompts.  Parse, request accuracy, and prompt-specific modal-vote
  accuracy are percentages.}
  \label{tab:generic-prompt-ablation}
\end{table}

\begin{figure}[H]
  \centering
  \includegraphics[width=0.92\textwidth]{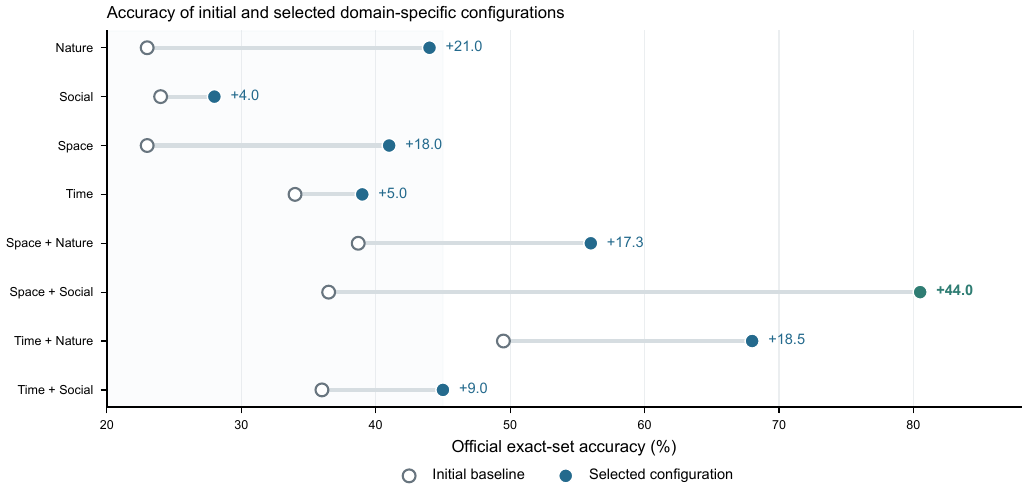}
  \caption{Comparison of domain-specific configurations. Open points show the
  initial baseline and filled points the selected configuration; labels
  report absolute accuracy differences.  All selected configurations score
  above their initial counterparts, with the largest difference on
  Space+Social.}
  \label{fig:route-ablation}
\end{figure}

Figure~\ref{fig:route-ablation} visualizes the route-wise differences, and
Table~\ref{tab:prompt-trajectory} compares successive prompt variants whose
accuracy did not improve monotonically.

\begin{table}[H]
  \centering
  \scriptsize
  \setlength{\tabcolsep}{5pt}
  \begin{tabular}{@{}lrrr@{}}
    \toprule
    Route & Initial prompt & Lower-scoring variant & Selected prompt \\
    \midrule
    Space & 24.0 (v1) & -- & 41.0 (v4) \\
    Space+Social & 39.0 (v1) & 35.0 (v3) & 80.5 (v4) \\
    Time & 33.0 (v1) & 21.0 (v2) & 39.0 (v3) \\
    Time+Social & \textbf{45.0} (v1) & 34.5 (v5) & \textbf{45.0} (v1) \\
    \bottomrule
  \end{tabular}
  \caption{Representative prompt comparisons from separate official domain
  evaluations. Labels v1--v5 identify successive prompt variants within each
  domain; variants can differ in multiple instructions.}
  \label{tab:prompt-trajectory}
\end{table}

\begin{figure}[H]
  \centering
  \includegraphics[width=\textwidth]{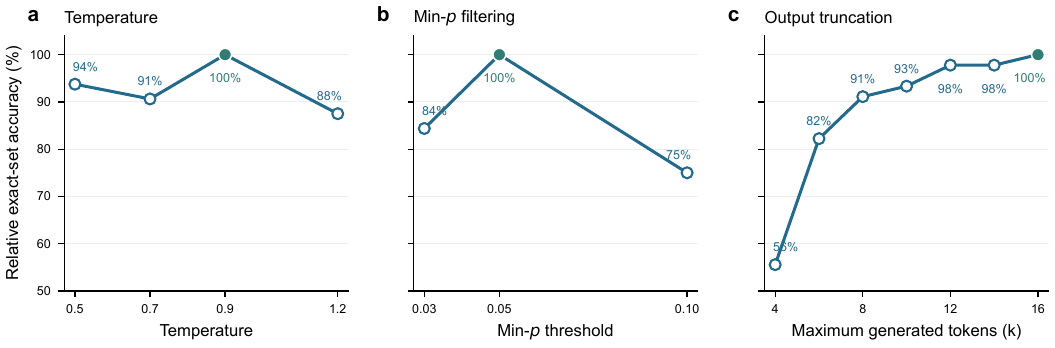}
  \caption{Sensitivity to sampling parameters. Each panel is
  normalized to its best observed exact-set accuracy (100\%).  \textbf{(a)}
  Temperature and \textbf{(b)} min-$p$ were measured on the 100-question
  official Space subset with a fixed domain-specific prompt; the unnormalized best
  is 32\%.  \textbf{(c)} Maximum generated tokens were evaluated by truncating
  and reparsing outputs from a preliminary 50-question experiment; the best vote
  accuracy is 90\%.}
  \label{fig:sensitivity}
\end{figure}

\paragraph{Sampling-parameter sensitivity.}
Figure~\ref{fig:sensitivity} summarizes two parameter sweeps on the official
Space subset and an output-truncation analysis. With the second spatial prompt
variant, temperature 0.9 and min-$p$ 0.05 attain the
best observed accuracy (32/100).  Relative to that point, the other
temperatures retain 88--94\% of the accuracy, while min-$p$ 0.03 and 0.10 retain
84\% and 75\%, respectively. In the output-truncation analysis, relative vote accuracy
rises from 56\% at 4k generated tokens to 98\% by 12k and 100\% at 16k.  Thus,
we selected moderate sampling parameters and a higher output-token limit.
Top-$k$ filtering was disabled throughout these experiments.

\FloatBarrier

\subsection{Test-Set Scaling}

Uniform answer-set voting improves monotonically as the sample budget grows,
but the gain from each doubling decreases.  Margin-triggered allocation reaches
77.3\% without assigning the maximum budget to every question.
Route-conditioned model pairing then raises the final accuracy to 79.4\%.

\begin{figure}[H]
  \centering
  \includegraphics[width=\textwidth]{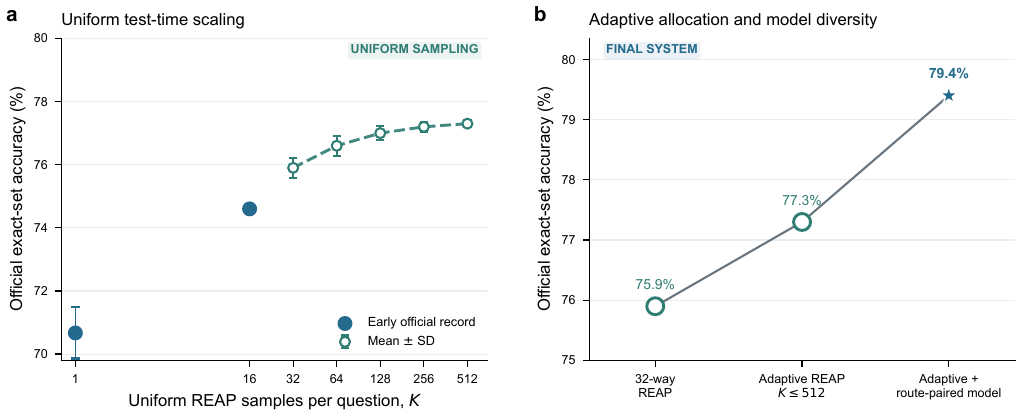}
  \caption{Official-test scaling and final system composition.
  \textbf{(a)} Uniform REAP voting improves with the number of sampled answer
  sets.  At $K=1$, error bars summarize available individual scores from the
  16 runs; at $K\geq32$, they summarize five
  repeated aggregates.  $K=16$ is one modal aggregate.
  \textbf{(b)} Adaptive REAP sampling and route-conditioned model pairing
  provide successive gains over the 32-output vote, reaching 79.4\%.}
  \label{fig:test-scaling}
\end{figure}

\begin{figure}[H]
  \centering
  \includegraphics[width=\textwidth]{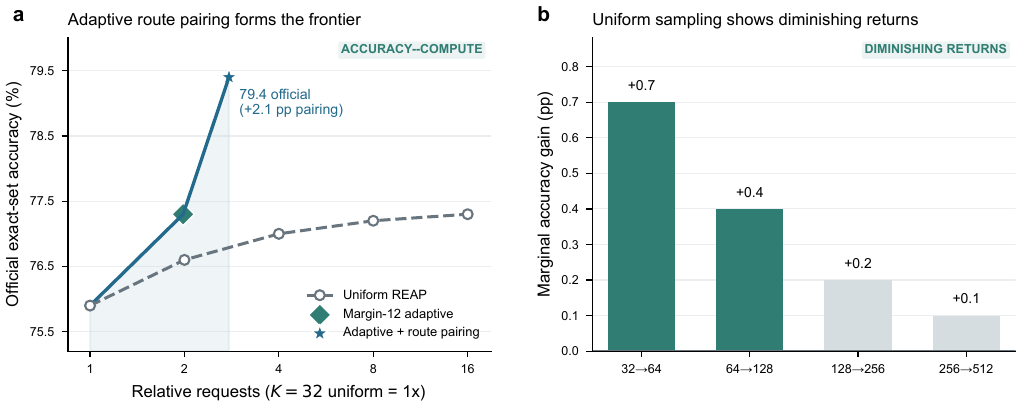}
  \caption{Accuracy--sampling-budget Pareto analysis. The number of sampled
  solutions is normalized by the uniform $K=32$ total (32,000), so $R=1$ is
  the baseline.  This proxy does not measure tokens or GPU time.
  \textbf{(a)} Uniform REAP scaling is compared with a
  margin-12 adaptive allocation and equal-count route-conditioned model pairing
  on 40\% of
  questions.  \textbf{(b)} Marginal gains from doubling uniform sampling.
  The decreasing increments motivate allocating additional computation to the
  low-margin subset instead of scaling every question uniformly.}
  \label{fig:pareto}
\end{figure}

Tables~\ref{tab:test-scaling}--\ref{tab:domain-scaling} detail the scaling
results at different sample budgets and by domain; Figure~\ref{fig:pareto} shows
the trade-off between accuracy and the number of sampled solutions.

\begin{table}[H]
  \centering
  \scriptsize
  \setlength{\tabcolsep}{8pt}
  \begin{tabular}{@{}rrr@{}}
    \toprule
    $K$ & Accuracy (\%) & SD \\
    \midrule
    1   & 70.67 & 0.81 \\
    16  & 74.6 & -- \\
    32  & 75.9 & 0.32 \\
    64  & 76.6 & 0.32 \\
    128 & 77.0 & 0.22 \\
    256 & 77.2 & 0.16 \\
    512 & 77.3 & 0.10 \\
    \bottomrule
  \end{tabular}
  \caption{Official-test accuracy under uniform answer-set voting as the
  number of sampled solutions increases.  At $K=1$, SD uses available
  individual scores from the 15 scored runs out of 16;
  at $K\geq32$, it summarizes five repeated aggregates.  $K=16$ is a single modal
  aggregate.}
  \label{tab:test-scaling}
\end{table}

\begin{table}[H]
  \centering
  \scriptsize
  \setlength{\tabcolsep}{5pt}
  \begin{tabular}{@{}rrrrrrrr@{}}
    \toprule
    Reached $K$ & 32 & 64 & 96 & 128 & 192 & 256 & 512 \\
    \midrule
    Active questions & 1000 & 387 & 226 & 152 & 50 & 19 & 1 \\
    Share (\%) & 100.0 & 38.7 & 22.6 & 15.2 & 5.0 & 1.9 & 0.1 \\
    \bottomrule
  \end{tabular}
  \caption{Questions requiring further sampling at selected sample counts.
  The average REAP budget of $K=63.4$ includes all 32-sample increments,
  including those not shown in the table.}
  \label{tab:adaptive-allocation}
\end{table}

\begin{table}[H]
  \centering
  \scriptsize
  \setlength{\tabcolsep}{5pt}
  \begin{tabular}{@{}lrrrrc@{}}
    \toprule
    Domain & $N$ & $K=32$ & Adaptive & Final & Pairing \\
    \midrule
    Nature & 100 & 69.0 & 71.0 & 77.0 & yes \\
    Social & 25 & 80.0 & 84.0 & 84.0 & no \\
    Space & 100 & 65.0 & 68.0 & 72.0 & yes \\
    Time & 100 & 71.0 & 71.0 & 71.0 & no \\
    \midrule
    Space+Nature & 75 & 77.3 & 78.7 & 78.7 & no \\
    Space+Social & 200 & 78.0 & 80.0 & 85.5 & yes \\
    Time+Nature & 200 & 80.0 & 80.0 & 80.0 & no \\
    Time+Social & 200 & 80.0 & 81.5 & 81.5 & no \\
    Weighted total & 1000 & 75.9 & 77.3 & \textbf{79.4} & three routes \\
    \bottomrule
  \end{tabular}
  \caption{Accuracy by domain for uniform REAP voting, adaptive REAP
  sampling, and the final route-conditioned model-pairing configuration.}
  \label{tab:domain-scaling}
\end{table}

\FloatBarrier

\subsection{Run Variability and Allocation}

\begin{figure}[H]
  \centering
  \includegraphics[width=\textwidth]{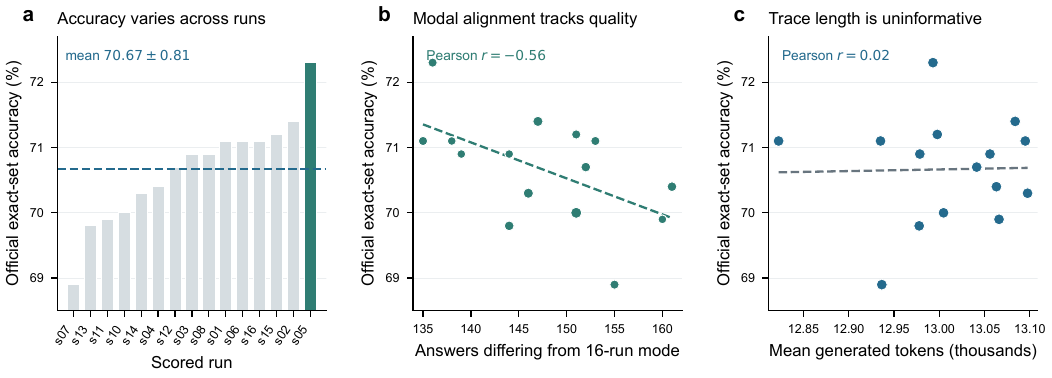}
  \caption{Run-to-run variability under a fixed reasoning procedure and decoding
  configuration.  \textbf{(a)} Available official exact-set scores from the
  16 runs; the dashed line is the mean, and the annotation
  reports the sample standard deviation.  \textbf{(b)} Official accuracy versus the number
  of answers differing from the 16-run modal answer.  Marker size indicates the
  number of empty parses.  \textbf{(c)} Official accuracy versus mean generated
  tokens.  Score-dependent panels omit the one run without individual official
  feedback, while the modal answer uses all 16
  outputs.}
  \label{fig:run-variability}
\end{figure}

\begin{figure}[H]
  \centering
  \includegraphics[width=\textwidth]{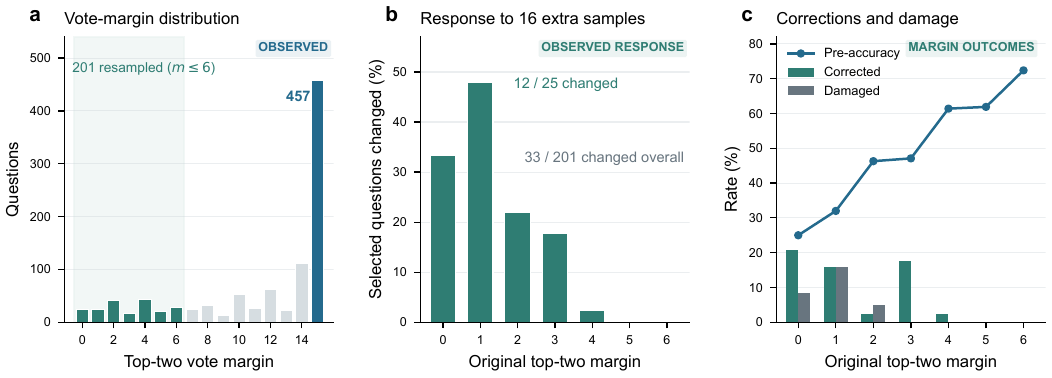}
  \caption{Vote margin identifies where added inference changes the
  prediction.  \textbf{(a)} Top-two margin distribution for the 16-output vote
  on 1,000 test questions. The additional-sampling experiment selects all 201 questions with
  margin at most six; 90 have margin at most two, whereas 457 are unanimous.
  \textbf{(b)} Fraction of selected questions whose modal answer changes after
  adding 16 samples, grouped by the original margin.  In total, 33 of 201
  answers change. \textbf{(c)} Pre-resampling
  accuracy, correction rate, and damage rate by margin.  Low-margin questions
  provide most corrections but also carry the greatest risk of changing an
  initially correct answer.}
  \label{fig:vote-allocation}
\end{figure}

Figure~\ref{fig:run-variability} examines repeated runs under a common
procedure and decoding configuration.  Accuracy varies despite similar output
lengths and parse coverage; agreement with the full 16-run modal answer provides
a clearer diagnostic.  Figure~\ref{fig:vote-allocation} then tests whether the
vote margin identifies where added samples alter the modal answer.

\FloatBarrier

\clearpage
{\normalsize
\bibliographystyle{ccl}
\bibliography{references}
}

\end{document}